\documentclass[runningheads]{llncs}

\usepackage{eccv}

\usepackage{eccvabbrv}

\usepackage{graphicx}
\usepackage{booktabs}

\usepackage[accsupp]{axessibility}  % Improves PDF readability for those with disabilities.
\usepackage{adjustbox}

\usepackage{hyperref}

\usepackage{orcidlink}

\begin{document}

% ---------------------------------------------------------------
% TODO REVIEW: Replace with your title
\title{Evaluating Image-Based Face and Eye Tracking
with Event Cameras} 

% TODO REVIEW: If the paper title is too long for the running head, you can set
% an abbreviated paper title here. If not, comment out.
\titlerunning{Face \& Eye Tracking with Event Cameras}

% TODO FINAL: Replace with your author list. 
% Include the authors' OCRID for the camera-ready version, if at all possible.
\author{Khadija Iddrisu\inst{1,3,4}\orcidlink{0009-0004-0008-5697} \and
Waseem Shariff\inst{2,3}\orcidlink{0000-0001-7298-9389} 
\and
Noel E. O'Connor\inst{4}\orcidlink{0000-0002-4033-9135} 
\and
Joseph Lemley \inst{1}\orcidlink{0000-0002-0595-2313}
\and
Suzanne Little\inst{1,4}\orcidlink{0000-0003-3281-3471}}

% TODO FINAL: Replace with an abbreviated list of authors.
\authorrunning{K.~Iddrisu et al.}
% First names are abbreviated in the running head.
% If there are more than two authors, 'et al.' is used.

% TODO FINAL: Replace with your institution list.
\institute{Dublin City University, Dublin, Ireland \\
\email{khadija.iddrisu2@mail.dcu.ie}\\
 \and
 University of Galway, Galway, Ireland\\
\email{w.shariff1@universityofgalway.ie}
\and
Drowsiness Team, FotoNation-Tobii, Galway,Ireland \and
\email{Joseph.Lemley@tobii.com}
 \and
Insight SFI Research Center for Data Analytics, Ireland\\
\email{\{Suzanne.Little,Noel.Oconnor\}@insight-centre.org}}
\maketitle

% \url{http://www.springer.com/gp/computer-science/lncs}

\begin{abstract}

Event Cameras, also known as Neuromorphic sensors, capture changes in local light intensity at the pixel level, producing asynchronously generated data termed ``events''. This distinct data format mitigates common issues observed in conventional cameras, like under-sampling  when capturing fast-moving objects, thereby preserving critical information that might otherwise be lost. However, leveraging this data often necessitates the development of specialized, handcrafted event representations that can integrate seamlessly with conventional Convolutional Neural Networks (CNNs), considering the unique attributes of event data. In this study, We evaluate event-based Face and Eye tracking. The core objective of our study is to showcase the viability of integrating conventional algorithms with event-based data, transformed into a frame format, while preserving the unique benefits of event cameras.  To validate our approach, we constructed a frame-based event dataset by simulating events between RGB frames derived from the publicly accessible Helen Dataset. We assess its utility for face and eye detection tasks through the application of GR-YOLO -- a pioneering technique derived from YOLOv3. This evaluation includes a comparative analysis with results derived from training the dataset with YOLOv8. Subsequently, the trained models were tested on real event streams from various iterations of Prophesee's event cameras and further evaluated on the Faces in Event Stream (FES) benchmark dataset. The models trained on our dataset shows a good prediction performance across all the datasets obtained for validation with the best results of a mean Average precision score of 0.91. Additionally, The models trained demonstrated robust performance on real event camera data under varying light conditions. 
\end{abstract}

\section{Introduction}
\label{intro}

Face and Eye Tracking are critical tasks in Computer Vision, with substantial applications in Healthcare, In-cabin Monitoring ~\cite{ryan2023real}, Attention Estimation ~\cite{sharma2022student, dhawan2022machine} and Human-Computer Interactions~\cite{bozomitu2019development}. This tracking technology is pivotal in detecting signs of fatigue, distraction, or impairment, necessitating a continuous stream of visual data, which is often unfeasible with traditional frame-based cameras. Other challenges include addressing scale variations as faces move closer or further from the camera, managing temporal dependencies between consecutive frames, accurately detecting faces under occlusions caused by rapid movements, and accounting for motion-induced shape deformations~\cite{yang2016wider, becattini2024neuromorphic}. 

Event cameras (ECs) on the other hand, respond to changes in brightness to produce a continuous stream of data of asynchronous nature called events. ECs capture high-speed motion with very low latency and minimal motion blur.  Events in an EC are represented by a stream of variable data points, each indicating a change in intensity at a specific pixel location at a given time.  An event is therefore represented as a tuple, $(x, y, t, p)$, where  $x, y$ are the spatial coordinates, $t$ is the timestamp, and  $p$ is the polarity. The polarity is used to indicate changes in pixel intensity. That is, for $p \in (0,1)$, $0$ indicates a decreasing change while 1 indicates an increasing change. ECs offer many advantages over traditional cameras including High Dynamic Range (140 dB versus 60 dB), High Temporal Resolution and Low Latency allowing them to capture and process visual information in real-time, with minimal delay making them ideal for applications that require fast, yet accurate visual feedback.

However, utilizing ECs for such tasks introduces unique challenges due to the distinct nature of events. It is a common practice in the domain of event-based vision to develop specialized algorithms with hand-crafted event representations to accommodate the use of this data. However, it is important to bridge the gap between this novel data type and the established paradigms of computer vision, which predominantly utilize Convolutional Neural Networks (CNNs) that process standard video frames~\cite{rebecq2019events}. In our work, we aim to highlight methods of representing events in a format accepted by existing deep learning approaches, specifically an image-based representation while preserving the unique advantages of event cameras. We do this by optimizing the frames generated during motion simulation from static images by maximizing the number of events produced and event-frame accumulation with Temporal Binary Representation (TBR)~\cite{innocenti2021temporal}. This study is inspired from the work of Ryan \etal~\cite{Ryan2021}. However, our methodology diverges from the previous approach in three fundamental aspects:

\begin{enumerate}
    \item We generate motion from the Helen Dataset by simulating planar motion from images placed in front of a camera in 6-Degrees of Freedom (DOF), as opposed to  cropping an image and shifting the crop along the $x$ and $y$ axes. 
    % This technique allows for the preservation of more information.

    \item Beyond employing an event simulator to convert RGB videos into events, we accumulate events into binary frames and aggregate these frames into a single frame. This process enhances the density and quality of the simulated frames derived from the original RGB frames, enriching better simulation of events.

    \item Finally, we carry out a detailed comparison, examining the performance of state-of-the-art models such as YOLOv8 and contrasting these results with those achieved by the GR-YOLO model~\cite{Ryan2021}. This evaluation is performed using simulated event datasets and real event datasets, thereby affirming the effectiveness and relevance of our proposed methods.
    
\end{enumerate}

\begin{figure}
    \centering
    \includegraphics[width=1\linewidth]{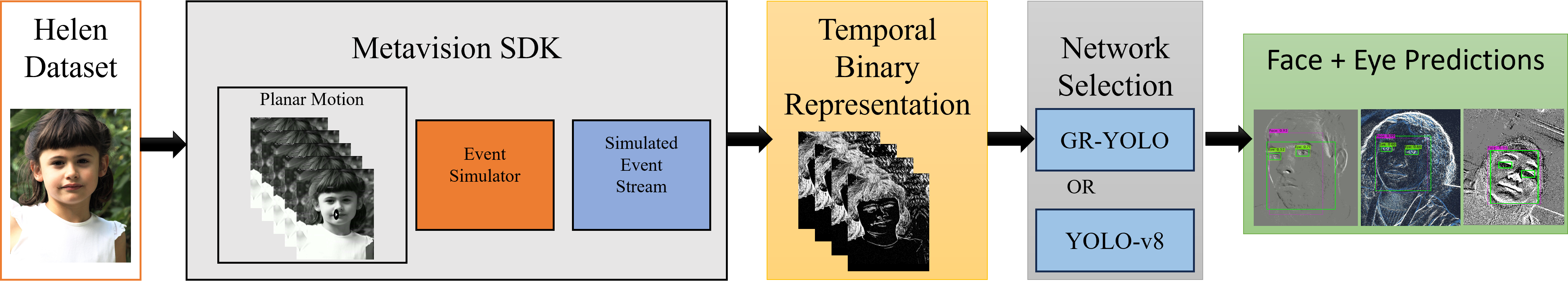}
   \caption{Overview of our proposed methodology}
    \label{fig:Fig1}
\end{figure}

Through this comparison, we aim to demonstrate the improvements our approach introduces to the domain and verify its usefulness in improving the adaptability and efficiency of deep learning models for processing the distinct data produced by event cameras. 
The paper is structured as follows: Literature Review, Dataset, Event Representation, Network Architecture, Training, Results and concluding remarks. We demonstrate the effectiveness of the proposed methodologies and show that training on these datasets generalizes well to real-world examples. An overview of the proposed methodology is illustrated in \cref{fig:Fig1}.

%%%%%%%%%%%%%%%%%%%%%%%%%%%%%%%%%%%%%%%%%%%%%%%%%%%%%%%%%%%%%%%%%%%%%%%%%%%%%%%%%%%

\section{Literature Review}

\subsection{Face Detection and Tracking}

Face tracking as a Facial Analysis task is critical, finding applications in neuroscience~\cite{wolf2021contribution,bortolon2016self}, automotive systems~\cite{dasgupta2013vision} \etc. The distinctive attributes of ECs, such as their low latency, facilitate the immediate reporting of scene changes with minimal delay. These characteristics among others have been demonstrated to offer promising applications in face analysis such as gaze estimation~\cite{li2024gaze,ryan2023real}, face pose alignment~\cite{savran2020face}, yawn detection ~\cite{kielty2023neuromorphic}, emotion recognition~\cite{becattini2022understanding, barchid2023spiking} \etc. However, these tasks have not achieved widespread attention within the domain of event-based vision. This is primarily due to the absence of event-based datasets that are readily applicable for face tracking.

Studies have revealed different approaches to the use of ECs for face tracking~\cite{becattini2024neuromorphic}.
The first works on ECs for face tracking leveraged different representations and algorithms that could fit these representations. Barua \etal~\cite{barua2016direct}, utilised a patch-based sparse dictionary generated from event data with the K-SVD algorithm, reconstructed high-intensity images from these streams and employed the Viola-Jones algorithm and random forest as a learning scheme for face detection. Lenz \etal~\cite{lenz2020event} introduced the first purely event-based approach for face detection and tracking, capitalizing on the high temporal resolution offered by an EC to identify the distinct signatures of eye blinks for face detection. Ramesh \etal~\cite{ramesh2020boosted} introduced a novel event-based feature learning method using kernelized correlation filters (KCF) within a boosting framework. Unlike previous works that relied on handcrafted feature descriptors, the proposed approach reformulates KCFs to learn face representations directly from data collected using ECs stemming from their capacity to sense asynchronous pixel-level brightness at a microsecond time-scale. Liu \etal~\cite{liu2022neurodfd} proposed a method for face detection by partitioning event streams into spatial-temporal volumes and  introduce a network comprising of a translation-invariant backbone, a Shift Feature Pyramid network (FPN), and a shift context module to handle the spatial-temporal nature of event data aiming for accurate and resource-efficient driver face detection.

In contrast to prior works, Current research have focused on representing events in a format understood by existing computer vision algorithms~\cite{rebecq2019events}. Such representations include:  Image Based~\cite{maqueda2018event,rudnev2021eventhands}, Surface Based~\cite{hu2021v2e,wang2020eventsr}, Graph-Based~\cite{baldwin2022time,zhang2024neuromorphic}, Voxel-Based~\cite{deng2022voxel,wang2023time} and Spike Based~\cite{cordone2021learning,wang2023spike}. Bissarinova \etal~\cite{bissarinova2023faces} acquired a large dataset of event streams with annotated landmarks and presents 12 models which they trained for face detection. Himmi \etal~\cite{himmi2024ms} proposes multi-spectral events, which incorporate multiple bands from the visible and near-infrared spectrum for face detection tasks over monochromatic events and traditional multi-spectral imaging resulting from RGB images that are simulated into events. 

%%%%%%%%%%%%%%%%%%%%%%%%%%%%%%%%%%%%%%%%%%%%%%%%%%%%%%%%%%%%%%%%%%%%%%%%%%%%%%%%%%%

\subsection{Eye tracking}

Prior research in eye tracking primarily relied on conventional cameras to identify and monitor eye movements, aiding in tasks such as activity recognition and attention assessment~\cite{singh2024robust,park2020towards}. However, recent attention has shifted towards ECs due to their asynchronous data presentation, eliminating the need for fixed frame rates. ECs offer low latency, enabling immediate event reporting with high temporal precision, making them particularly suitable for tracking eyes in scenarios involving rapid motion.

Foundational works by Angelopoulos \etal~\cite{angelopoulos2020event} used traditional image processing and statistical methods for segmenting eye regions effectively. Building on this, Feng \etal~\cite{feng2022real} introduced an innovative Auto ROI algorithm that dynamically predicts eye ROIs to enhance tracking efficiency. Further, Li \etal~\cite{li2023track} adopted an event-driven approach, processing event streams into frames and employing a low-latency CNN with an event-based ROI system for accurate pupil detection.

In contrast, a series of studies leveraged neural networks to address the unique challenges posed by sparse event data. Chen \etal ~\cite{chen20233et} proposed a Change-Based Convolutional Long Short-Term Memory (CB-ConvLSTM) model for precise pupil tracking. Bonazzi \etal~\cite{bonazzi2023low} utilized a Spiking Neural Network (SNN) trained directly on event data, and Yang \etal~\cite{yang2023high} applied a U-Net based network alongside frame interpolation techniques to produce high frame rate videos from event streams for detailed pupil segmentation.

Recent studies have introduced more specialized techniques. Stoffregen \etal~\cite{stoffregen2022event} focused on detecting corneal glint using a coded differential lighting system to improve specular reflection detection in a purely event-based manner. Zhang \etal~\cite{zhang2024swift} proposed event-based frame interpolation, and a suite of modules for feature extraction and temporal feature fusion, particularly effective during blinking motions. Lastly, Zhao \etal~\cite{zhao2024ev} employed polynomial regression to localize pupil centroids and accurately identify the Point of Gaze from event streams.

Quite recently, a number of studies have emerged on the use of ECs for eye tracking as a result of the introduction of the Event-based Eye Tracking(EET) challenge~\cite{wang2024event}. The CVPR AI in Streaming (AIS) EET Challenge on Kaggle focused on the use of data from ECs for eye tracking. In our study, we do not review individual studies due to constraints on length. It is however worth noting that most participants utilised frame based representations with already existing deep learning algorithms such as CNNs,GRU and LSTMs, demonstrating the significance of our evaluation. These developments showcase the diversification and progression of techniques in event-based eye tracking, from traditional methodologies to novel approaches tailored to exploit the full capabilities of ECs. A more comprehensive survey of the solutions presented is provided in the associated survey of the challenge~\cite{wang2024event}.

%%%%%%%%%%%%%%%%%%%%%%%%%%%%%%%%%%%%%%%%%%%%%%%%%%%%%%%%%%%%%%%%%%%%%

\section{Dataset}

The availability of datasets in event-based vision, particularly for face and eye tracking, remains limited. Most research utilizes locally sourced or simulated datasets~\cite{shariff2023neuromorphic,berlincioni2024neuromorphic}. Publicly available options like the Neuromorphic Event-based Facial Expression Recognition dataset~\cite{berlincioni2023neuromorphic} offer RGB and event streams for facial expression recognition but lack face and eye location annotations of the event stream. Directly applicable datasets for face and eye tracking are scarce, with the recent Faces in Event Stream (FES) dataset~\cite{bissarinova2023faces} being a notable exception. Despite providing event streams and bounding boxes, FES lacks eye annotations, necessitating manual annotation efforts. 

A promising solution involves using event simulators. For instance, Ryan \etal~\cite{Ryan2021} developed the \emph{Neuro-morphic Helen} synthetic dataset by converting the Helen Facial Landmarks Dataset, which comprises 2,300 internet-sourced images with landmark annotations, into an event format. This conversion involved simulating camera motion across the images and applying random augmentations before feeding the resulting videos into an event simulator that transforms RGB videos into events~\cite{hu2021v2e}. 
A main advantage in the use of the Helen dataset for this task is the different facial features present in the dataset that enables us to train a model robust to a wide variety of facial features and occlusions such as glasses, headwear, \etc.

Reproducing the Neuro-morphic Helen Dataset reveals challenges in how Event Simulators mimic real-world event generation. These simulators often rely on discrete time intervals to model events, deviating from the continuous flow of real-world dynamics~\cite{becattini2024neuromorphic, berlincioni2024neuromorphic}. To address this issue, more sophisticated algorithms are required that can approximate continuous real-world processes while balancing computational feasibility.
To generate motion similar to that of real Prophesee ECs~\cite{Prophesee}, we utilized the \texttt{PlanarMotionStream} object of the Metavision software suite \cite{Metavision}. This software simulation operates in a manner that is similar to the way  Prophesee event cameras function, ensuring that the events generated are similar to those obtained from real Prophesee ECs. The \texttt{PlanarMotionStream} class creates a simulated stream of images that depict how a static picture would appear if observed through a camera undergoing planar motion in front of it.

\texttt{PlanarMotionStream}  is designed to create a realistic simulation of 6-DOF (Degrees of Freedom) motion for a given image. The class generates a sequence of frames, each representing the image as viewed from a slightly different camera pose. This is achieved through the application of homographic transformations that preserve straight lines and are widely used for tasks such as image stitching and perspective correction. The class is initialized with  several parameters with the ones of utmost importance being; \texttt{infinite} - a boolean flag determining the border handling method (mirrored or constant), $
\texttt{pause\_probability}$ - probability that the camera motion will pause at each frame,  $\texttt{max\_frames}$ - the maximum number of frames to stream, and a method -
$\texttt{get\_relative\_homography}$ which computes the homography between two camera poses. Given a specific time step, it retrieves the rotation and translation vectors for the specified time step and the current iteration, and then calculates the relative homography using these vectors. This method is useful for understanding the transformation between any two frames in the sequence.

\begin{figure}[ht]
    \centering
    \includegraphics[width=1\linewidth]{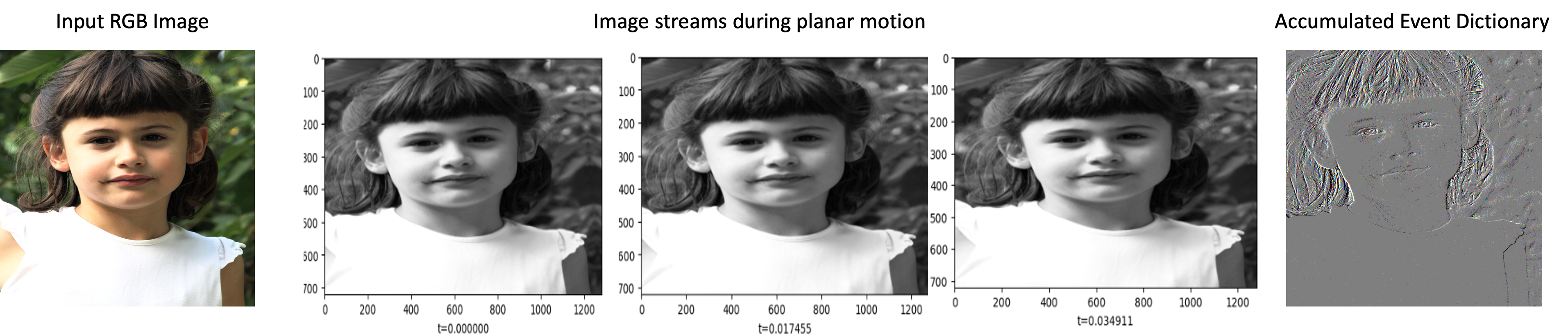}
   \caption{A sample video showcasing motion derived from an RGB image, transformed into events and then rebuilt into an event frame. From the left: original RGB image, 3 frames showing generated motion and compiled event frame.}
    \label{fig:Fig2}
\end{figure}

We  set  $\texttt{pause\_probability} = 0.5$ and $\texttt{max\_frames}=  100 $ contributing to a more realistic simulation. This allows a smooth motion that does not lead to a rapid shift in faces from the original position of the facial features in the RGB image. Smoother transitions maintain the continuity of facial features, which is crucial for accurate event representation. This process aids in solving the problem of under-sampling as more interpolated frames are generated to produce a continuous stream of events that accounts for every pixel value in the final frames used for training. Once the motion-generated videos based on a fixed frame rate have been created, we use the event simulator object from the same SDK to convert these videos into event dictionaries for further preprocessing. An example is seen in \cref{fig:Fig2}.

%%%%%%%%%%%%%%%%%%%%%%%%%%%%%%%%%%%%%%%%%%%%%%%%%%%%%%%%%%%%%%%%%%%%%%%%%%%%%%%%%%%

\section{Event Representation}\label{sec:Event Representation}

A typical characteristic of events is the structural differences between the data produced and that used as input for Convolutional Neural Networks (CNNs). For object tracking there is a need to transform event data into a representation accepted by CNNs. In recent years, there have been several hand crafted event-representations proposed. In this study, instead of using hand-crafted temporal representations, we employed an image-based representation to enable training with existing object detection models. This methodology involves converting the simulated event streams into two-dimensional frames, which are directly interpretable by CNNs. The approach also focuses on preserving key information from the events, such as polarity, timestamps, and in some cases, event count. 

\begin{figure}
	\centering
	\begin{subfigure}{0.33\linewidth}
		\includegraphics[width=\linewidth]{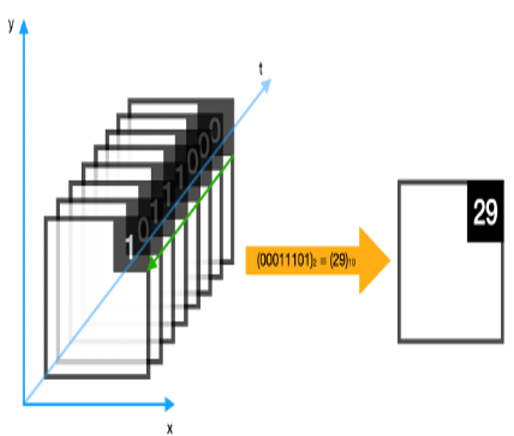}
	 \caption{Events in a single frame for a time window}\label{fig3:a}
	\end{subfigure}
	\begin{subfigure}{0.32\linewidth}
		\includegraphics[width=\linewidth]{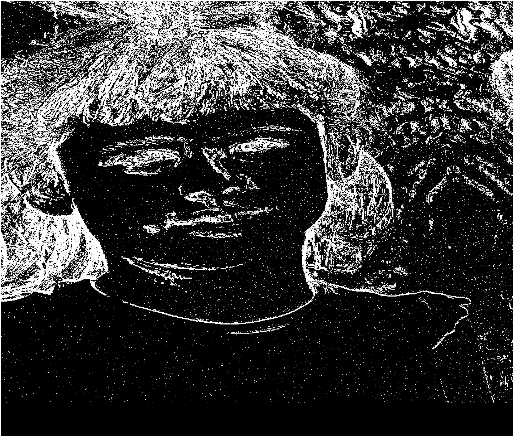}
		 \caption{Example binary frame from TBR}\label{fig3:b}
	\end{subfigure}
	\begin{subfigure}{0.32\linewidth}
	        \includegraphics[width=\linewidth]{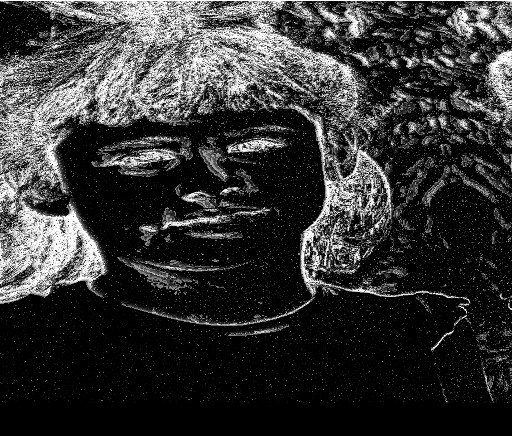}
  \caption{Accumulated frame from binary frames}\label{fig3:c}
         \end{subfigure}
	\label{fig:Fig3}
 \caption{Event Representation Procedure: The value
in position $(x, y)$ is obtained as $b^i
_{x,y}= \mathbf{1}(x, y)$, where $\mathbf{1}(x, y)$
is an indicator function returning 1 if an event is present in
position ($x$, $y$) and 0 otherwise (image from~\cite{innocenti2021temporal})}
\end{figure}

We leveraged the Temporal Binary Representation (TBR) method of aggregating events~\cite{innocenti2021temporal}. TBR employs a binary method for processing events in pixels over a fixed time window, $\Delta t$. A binary representation is created for each pixel indicating the presence $(1)$ or absence $(0)$ of an event during the accumulation time. Finally, these binary values are stacked into a tensor, with each pixel represented as a binary string. This string is then converted into a decimal number, allowing the representation of $N$ consecutive accumulation times in a single frame without information loss, as shown in \cref{fig3:a}. The frame is normalized by dividing its values by $N$. \cref{fig3:b} represents 1 binary frame from several binary frames generated and \cref{fig3:c} shows the resulting accumulated frames. This allows us to encode temporal information
directly into pixel values, which are then interpreted by CNNs.

TBR offers significant advantages over traditional event aggregation methods. By reducing the memory requirement by a factor of $N$, it minimizes the data processed by computer vision algorithms. This efficiency enables applications under time constraints and allows for capturing events at finer temporal scales without increasing the number of frames. 

%%%%%%%%%%%%%%%%%%%%%%%%%%%%%%%%%%%%%%%%%%%%%%%%%%%%%%%%%%%%%%%%%%%%%%%%%%%%%%%%%%%

\section{Network Architecture}
This section outlines the network architectures employed in the study to assess the efficacy of traditional Face and Eye tracking techniques compared to the state-of-the-art YOLOv8 based method.  

%%%%%%%%%%%%%%%%%%%%%%%%%%%%%%%%%%%%%%%%%%%%%%%%%%%%%%%%%%%%%%%%%%%%%%%%%%%%%%%%%%%

\subsection{GR-YOLOv3}

The research methodology closely aligns with the experimental framework proposed by Ryan \etal~\cite{Ryan2021} and utilizes the custom GR-YOLO architecture. A notable deviation in our approach is eschewing voxel grids, typically employed in discrete time bins, and instead leveraging frame-based representations. This decision simplifies the model's interface with existing architectures and enhances adaptability and applicability in real-world scenarios. The GR-YOLO architecture, shown in ~\cref{tab:1}, integrates the YOLOv3 Tiny model with the addition of a Gated Recurrent Unit (GRU). The network comprises several key layers, each contributing uniquely to the effectiveness of the model.

\begin{table}[ht]
\caption{GR-YOLO Architecture: layers, filters, inputs \& output dimensions}
\centering
\label{tab:1}
\begin{tabular}{|l|l|l|l|l|l|}
\hline
\textbf{Layer} & \textbf{Type} & \textbf{Filter} & \textbf{Kernel/Stride} & \textbf{Input} & \textbf{Output} \\ \hline
0              & Conv          & 16              & 3/1                    & 256 x 256 x 1  & 256 x 256 x 16 \\ 

1              & Maxpool          &               & 2/2                   & 256 x 256 x 16  & 128 x 128 x 16 \\ 

2            & Conv          & 32              & 3/1                    &  128 x 128 x 16 &  128 x 128 x 32 \\ 

3             & Maxpool          &               & 2/2                    & 128 x 128 x 32  & 64 x 64 x 32 \\ 

4              & Conv          & 64             & 3/1                    & 64 x 64 x 32  & 64 x 64 x 64 \\ 

5              & Maxpool          &              & 2/2                    &  64 x 64 x 64  &  32 x 32 x 64 \\ 

6              & Conv          & 128              & 3/1                    & 32 x 32 x 64  & 32 x 32 x 128 \\

7              & Maxpool          &              & 2/2                   & 32 x 32 x 128  & 16 x 16 x 128 \\

8            & Conv          & 256             & 3/1                    & 16 x 16 x 128  & 16 x 16 x 256 \\ 

9             & Maxpool          &               & 2/2                   & 16 x 16 x 256   & 8 x 8 x 256  \\ 

10              & Conv          & 512              & 3/1                    & 8 x 8 x 256  & 8 x 8 x 512\\ 

11              & Maxpool          &              & 2/2                   &8 x 8 x 512  & 8 x 8 x 512 \\ 

12              & Conv          & 1024              & 3/1                    &  8 x 8 x 512 &  8 x 8 x 1024\\ 
13              & Conv          & 256              & 1/1                    &  8 x 8 x 1024 &  8 x 8 x 256 \\ 
14            & \textbf{GRU}          & 256              & 3/1                    & 8 x 8 x 256 & 8 x 8 x 256 \\ 

15            & Conv          & 512             & 3/1                    &  8 x 8 x 256  &  8 x 8 x 512 \\ 

16              & Conv          & 21              & 1/1                    & 8 x 8 x 512  & 8 x 8 x 21 \\ 

17              & \textbf{YOLO}          &              &                    & 8 x 8 x 21  & 192 x 7 \\ 

18              & \textbf{ROUTE 14}          &               &                     &   & 8 x 8 x 256 \\ 
19             & Conv          & 128              & 1/1                    & 8 x 8 x 256  & 8 x 8 x 128 \\ 
20            & \textbf{Up-Sampling}          &               &                     & 8 x 8 x 128  & 16 x 16 x 128 \\ 

21             & \textbf{ROUTE 20 8}          &               &                     &  & 16 x 16 x 384 \\ 

22             & Conv          & 256              & 3/1                    &16 x 16 x 384  &16 x 16 x 256 \\ 

23              & Conv          & 21              & 3/1                    &16 x 16 x 256 & 16 x 16 x 21 \\
24              & \textbf{YOLO}          &               &                 &16 x 16 x 21 & 768 x 7 \\
\hline

\end{tabular}
\end{table}

The GRU is particularly beneficial in scenarios where sequences of frames are involved. By retaining information from previous frames, the GRU enables the model to understand motion and changes over time, leading to more accurate and consistent tracking of faces and eyes. This backward propagation of information significantly enhances the model's performance compared to the Tiny-YOLOv3 model without a GRU. YOLO Heads are responsible for predicting bounding boxes and class probabilities for detected objects. Having two YOLO heads implies that the model can process different scales of detection simultaneously, improving its accuracy and robustness in detecting objects of various sizes. 

\subsection{YOLOv8}
YOLOv8~\cite{yolov8_ultralytics} offers enhanced accuracy and speed, making it suitable for real-time applications. YOLOv8  is characterized by its Fully Convolutional Architecture, which allows for efficient processing of images. As there is no published paper for this model, inference and understanding of the research methodology are achieved mainly by the published code. One of the critical improvements in YOLOv8 over its predecessors is its enhanced backbone network. This backbone is responsible for feature extraction and is more adept at handling small and occluded objects, which is a common challenge in face and eye tracking scenarios. Another improvement worth noting is the elimination of anchor boxes. These have largely been a part of earlier models as they represent the distribution and center-points of bounding boxes. This has resulted in the reduction of the number of predicted boxes, consequently speeding up Non-Maximum Suppression (NMS).

The architecture of YOLOv8 is an enhancement of YOLOv5 where the first $6 \times 6$ convolutional stem is replaced by a $3 \times 3$. CBS blocks consisting of Convolutional, Batchnorm and SILU are used to replace the main building block and C2f. C2f represents two $3 \times 3$ convolutions with a residual connection which accepts outputs from the bottleneck and concatenates it. The bottleneck is the same as in YOLOv5 with changes made to the size of the first convolutional layer from $1 \times 1$ to $3 \times 3$. YOLOv8 also includes significant changes during training. Augmentation is applied to every image at each epoch, one of which, the mosaic augmentation, allows the model to learn objects in new locations making it invariant to changes at each epoch~\cite{sohan2024review}.

%%%%%%%%%%%%%%%%%%%%%%%%%Training%%%%%%%%%%%%%%%%%%%%%%%%%%%%%%%%%%%%%%

\subsection{Training}

In our study, we utilised the event frames generated from the methodology reviewed in \cref{sec:Event Representation} to train the baseline model (GR-YOLO) and YOLOv8. The dataset, comprising 2,330 accumulated frames, was divided into a training and validation set with a ratio of 80:20 respectively. The training process was implemented in PyTorch and trained on an NVIDIA GeForce RTX 2080 Ti GPU. An AdamW optimizer was utilised in both models to achieve the highest performance with a learning rate of  $1 \times 10 ^{-3}$ and a weight decay of  $1 \times 10^{-3}$. With trained using the Mean Squared Error (MSE) as the loss function, with the loss being calculated across both detection layers of the GR-YOLO algorithm.

% ###################### Results ############################

\section{Experiments and Evaluation}
In this section, we present the results for both our baseline model and YOLOv8. To assess the efficacy of our trained models on our dataset and to demonstrate that the Frame-Based Representation of events from TBR effectively generalizes to real event camera datasets, we will conduct a comprehensive evaluation. This assessment will include both quantitative and qualitative analysis of the performance of our models when applied to real event camera datasets.

%%%%%%%%%%%%%%%%%%%%%%%%%%%%%%%%%%%%%%%%%%%%%%%%%%%%%%%%%%%%%%%%%%%%%%%%%%%%%%%%%%%

\subsection{Quantitative Results: Synthetic Data Evaluation}

We assess the effectiveness and suitability of our dataset for face and eye tracking by comparing the outcomes of testing three iterations of YOLO models on the test set of our synthetic data. This comparison will be against the performance metrics reported by Ryan \etal~\cite{Ryan2021} when they applied GR-YOLO to their synthetic dataset. A more appropriate and direct comparison would have entailed evaluating the performance of our trained models on the synthetic datasets created by the authors. This direct comparison would allow us to evaluate the difference in performance of our proposed method of data generation as well as our method of event representation. However, this was not feasible due to our lack of access to their dataset. Our analysis, detailed in \cref{table:2}, evaluates our frame-based methodology for face and eye tracking with ECs against the voxel grid approach used in the referenced study. This comparison allows us to critically examine the relative strengths of each approach using the results from the respective synthetic datasets 

\begin{table}[ht!]
\centering
\caption{Comparison of performance metrics for three YOLO models GR-YOLO (voxel grids~\cite{Ryan2021} and frames(ours)), YOLOv3 and YOLOv8 on Synthetic dataset.}
\begin{adjustbox}{width=\textwidth, center}
\begin{tabular}{|l|l|l|l|l|}
\hline
\textbf{Metric} & \textbf{GR-YOLO (Voxel grids)} & \textbf{GR-YOLO (frames)} & \textbf{YOLOv3} & \textbf{YOLOv8} \\ \hline
Mean Average Precision  & 0.95  & 0.91 & 0.81 & 0.94 \\
Mean Squared Error & 0.71     & 0.82  & 1.33  & 0.74  \\ 
Average Recall & 0.95    & 0.88 & 0.81   & 0.91  \\
F1-Score & --  & 0.86  & 0.81       & 0.91  \\  \hline
\end{tabular}
\end{adjustbox}

\label{table:2}
\end{table}

Ryan \etal~\cite{Ryan2021} demonstrated a robust performance with the top mean Average Precision (mAP) closely followed by YOLOv8. It is also observed that GR-YOLO still results in significant performance when compared to YOLOv3 without GRU, suggesting that this adaptation of YOLOv3 is applicable to other representations aside from voxel grids. The lowest error is observed in YOLOv8, which is close to the results obtained by Ryan \etal. YOLOv3 has the highest MSE with 1.33, which indicates significant inaccuracies in its predictions. The average recall metric, which evaluates the model's ability to detect all relevant instances, indicates a superior capacity to identify relevant objects in YOLOv8. The recall rates for GR-YOLOv3 and YOLOv3 are lower. Despite the absence of the F1-Score data, the prevailing metrics suggest that Ryan \etal's~\cite{Ryan2021} model surpasses the others in terms of precision, accuracy, and recall. This is quite comprehensive as voxel grids offer a superior representation. This could also be a result of other factors such as the type of augmentations used during synthetic data generation which we did not have information on.

Nonetheless, the frame-based approach we presented demonstrates excellent performance, with YOLOv8 leading in terms of error minimization and localization accuracy. Consequently, we have successfully demonstrated that face and eye tracking can be achieved using a frame-based representation of events. This method generates robust results, making it a preferable choice for applications requiring high reliability and low computational demand. Additionally, it addresses the issues of under-sampling present in traditional RGB cameras for this task.

%%%%%%%%%%%%%%%%%%%%%%%%%%%%%%%%%%%%%%%%%%%%%%%%%%%%%%%%%%%%%%%%%%%%%%%%%%%%%%%%

\subsection{Quantitative Results: Evaluation on Real Event Camera Data}

In the subsequent phases of evaluation, we verify the suitability of the models trained on our synthetic dataset against diverse data datasets gathered by different models of real ECs. To begin with, we test the trained models against the largest and publicly available event-based dataset for face tracking, the recent Faces in Event Stream (FES) dataset~\cite{bissarinova2023faces}. This dataset serves as the most comparable benchmark in event-based vision, comprising 1.6 million event streams captured with a Prophesee PPS3MVCD EC, incorporating face bounding box annotations.  This data was obtained by sending a formal request to the authors and subsequently, we were given the permission to access and utilise the data for this study. Given that our focus includes eye tracking, we manually annotated the eyes of selected test subjects within FES to facilitate a balanced comparison.

Additionally, we evaluate the performance of our models using naturalistic driving data recorded by Ryan \etal~\cite{Ryan2021}. The dataset, which was not publicly accessible, was obtained for our study by requesting its use as a benchmark. Likewise, permission to utilize this data for our research was granted by the authors. The direct comparison with this data enables us to rigorously evaluate our frame-based methodology for face and eye tracking with ECs against the Voxel Grid strategy employed in the cited study. 

In our final phase of evaluation, we collected sample data using an EVK4 EC, which captured a spectrum of head movements from minimal to extremely rapid.  The task of manually annotating bounding box labels for faces and eyes in each frame for the last 2 datasets, though time-consuming, was imperative. This detailed annotation process was critical not only for deriving precise results but also for illustrating the effectiveness of our models across diverse datasets and with varying event cameras. All datasets employed in our testing phase were excluded from the training to ensure an unbiased evaluation of the performance of our models.

\begin{table}[ht!]
\centering
\caption{Comparison of performance metrics of baseline model and YOLOv8 on real event camera datasets.}
\begin{adjustbox}{width=\textwidth, center}
\begin{tabular}{| *{7}{c|} }
\hline
 & \multicolumn{2}{c|} {\textbf{FES~\cite{bissarinova2023faces} }}
            & \multicolumn{2}{c|}{\textbf{Ryan \etal ~\cite{Ryan2021}}}
                    & \multicolumn{2}{c|}{\textbf{Local data}} \\
\hline
% \textbf{Metric} & \textbf{FES~\cite{bissarinova2023faces} } & & \textbf{Ryan \etal ~\cite{Ryan2021}} & &\textbf{Local data} &\\ \hline

\textbf{Metric}&GR-YOLOv3&YOLOv8&GR-YOLOv3&YOLOv8&GR-YOLOv3&YOLOv8\\\hline

Mean Average Precision (All)  & 0.70 &0.83  & \textbf{0.92} &\textbf{0.97}  & 0.88 & 0.90 \\
Mean Average Precision (Face) & 0.94 & 0.98 & \textbf{0.99} &  \textbf{0.99} & 0.96 &0.97\\
Mean Average Precision (Eye) & 0.45  &0.67  & \textbf{0.85} &\textbf{0.91} & 0.81 &  0.84 \\
Precision & 0.75  & 0.84  & \textbf{0.89} &\textbf{0.97}  & 0.86  &0.90  \\ 
Recall & 0.67  &0.82  & 0.88 & \textbf{0.92} & \textbf{0.89}   &0.88 \\
  
\hline
\end{tabular}
\end{adjustbox}

\label{table:3}
\end{table}

\cref{table:3} contains the validation outcomes of our baseline model (GR-YOLO) and YOLOv8 across the test datasets employing the same evaluation metrics for a consistent and comparative analysis. 
This facilitates a clear understanding of each model's effectiveness and efficiency in real-world application scenarios. All the datasets employed in our testing phase were excluded from the training to ensure an unbiased evaluation of the performance of our models.

The GR-YOLO model exhibits a robust performance in face detection, particularly when validated with the dataset of Ryan \etal~\cite{Ryan2021}. This was anticipated as the motion simulating approach employed was based upon the operation of a prophesee EC. Therefore, the data obtained from the prophesee EC should yield highly accurate prediction outcomes. The FES dataset shows a decrease in validation results across all metrics. Though it shows good performance in face mAP, this is half what is shown in eye mAP. As the FES dataset is more focused on detecting faces than eyes, some event streams do not highlight the eyes. Therefore, frames from these events were not annotated during manual annotation to avoid assuming the positions of eyes. Thus lower results in face mAP are to be expected, which affects the overall performance of other metrics.  

Generally across all datasets tested, the efficacy of GR-YOLO in eye detection exhibits a greater variability, accompanied by a significantly lower mAP. This highlights a challenge often encountered when using YOLOv3 for detecting smaller or more detailed objects like eyes. This is more apparent in ECs where the motion is relatively low, and as such, only a few events or no events are generated in eye regions. The overall performance measured by mAP across all datasets is satisfactory, indicating a comprehensive capability to detect faces and eyes from diverse sensors. Precision and recall metrics also support this conclusion. This demonstrates that GR-YOLO not only accurately identifies objects, but also performs consistently across datasets.

In contrast, YOLOv8 shows a noticeable improvement in performance across all metrics and datasets compared with GR-YOLOv3. The mAP for faces remains high, with a slight improvement in the FES Dataset and consistent performance in Ryan \etal. The eye detection capability  significantly improved, as evidenced by the increase in mAP for eyes in the FES Dataset. This enhancement indicates that YOLOv8 better handles the intricacies of eye detection across different sensors. The overall metric evaluation across all objects also sees an increase highlighting the superior general object detection capability of YOLOv8 over GR-YOLOv3 and underscoring the model's accuracy and consistency in object detection across varied conditions.

% \input{Event_Evaluation}

%%%%%%%%%%%%%%%%%%%%%%%%%%%%%%%%%%%%%%%%%%%%%%%%%%%%%%%%%%%%%%%%%%%%%%%%%%%%%%%%%%%

\subsection{Qualitative Results}
\begin{figure}

	\centering
	\begin{subfigure}{0.3\linewidth}
		\includegraphics[width=3.5cm, height = 3cm]{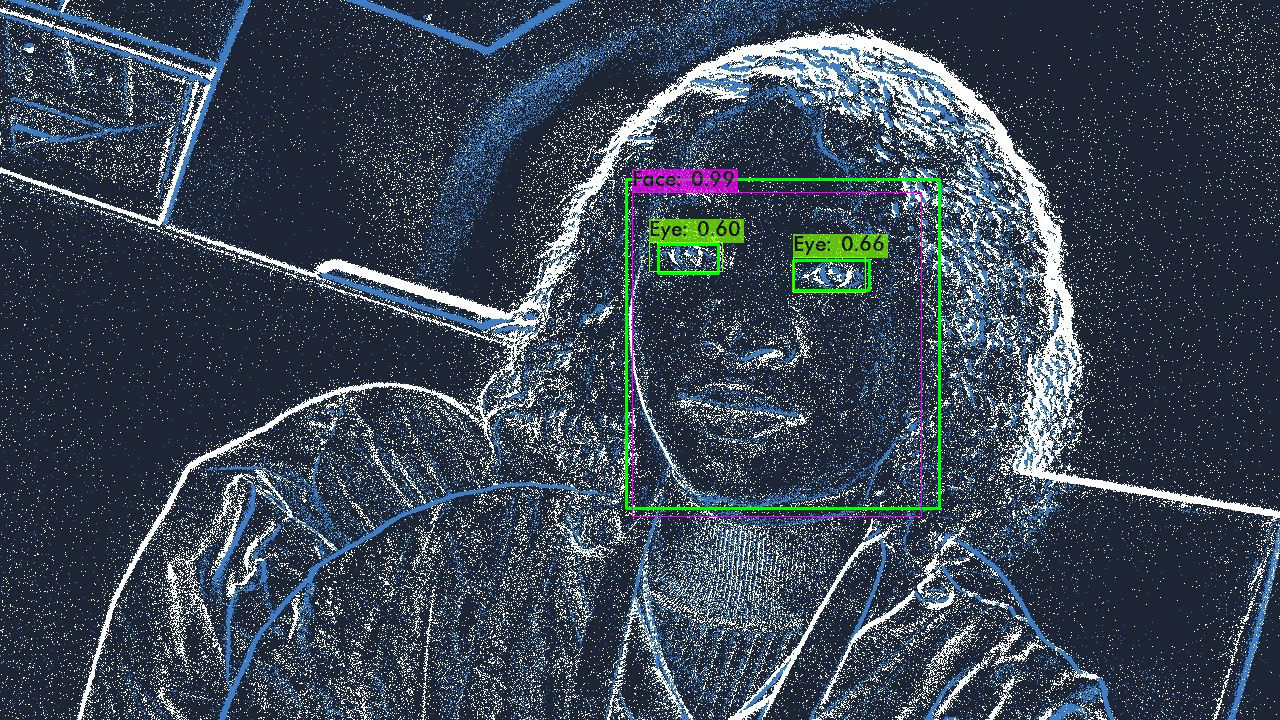}
		\caption{Local data}
		\label{fig4:a}
	\end{subfigure}
	\begin{subfigure}{0.3\linewidth}
		\includegraphics[width=3.5cm, height = 3cm]{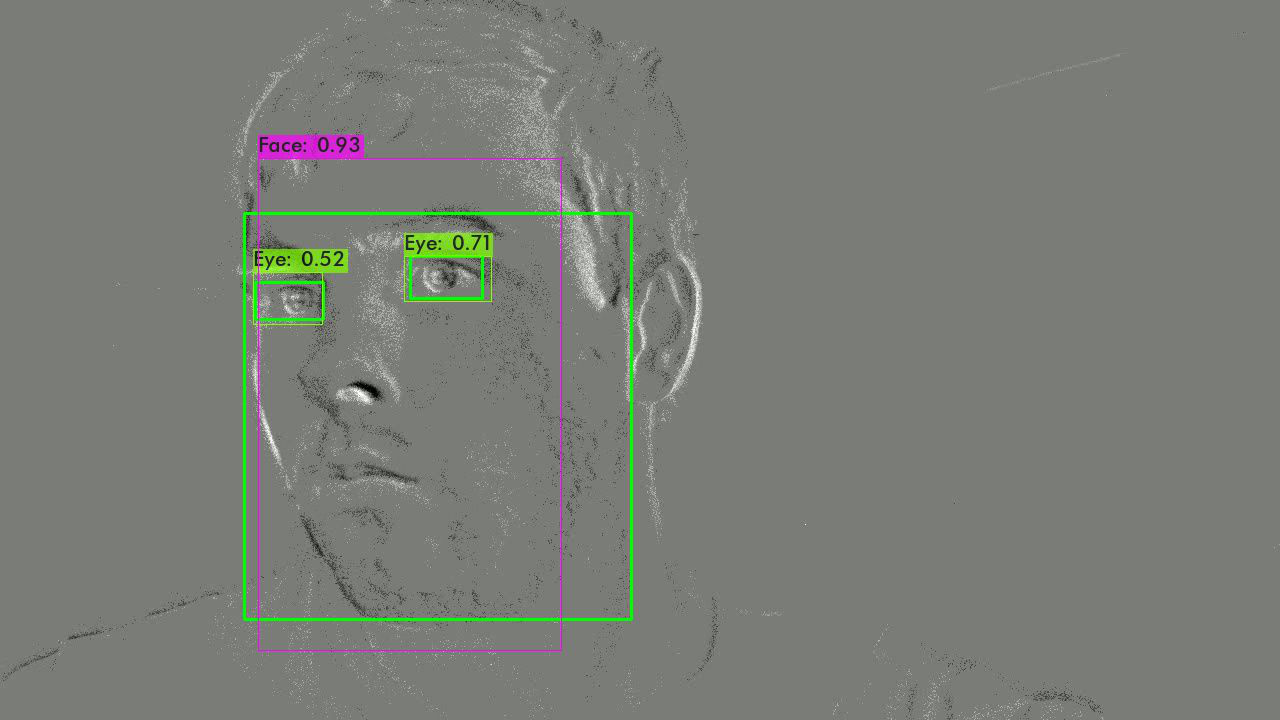}
		\caption{Data by Ryan et al ~\cite{Ryan2021}}
		\label{fig4:b}
	\end{subfigure}
	\begin{subfigure}{0.3\linewidth}
	        \includegraphics[width=3.5cm, height = 3cm]{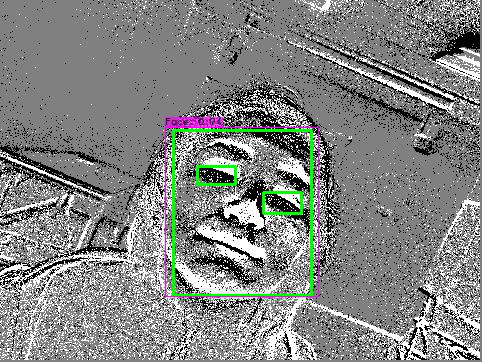}
	        \caption{FES data}
	        \label{fig4:c}
         \end{subfigure}
	\caption{Prediction Performance of GR-YOLOv3.}
	\label{fig3:Test_V3}
\end{figure}

\begin{figure}

	\centering
	\begin{subfigure}{0.3\linewidth}
		\includegraphics[width=3.5cm, height = 3cm]{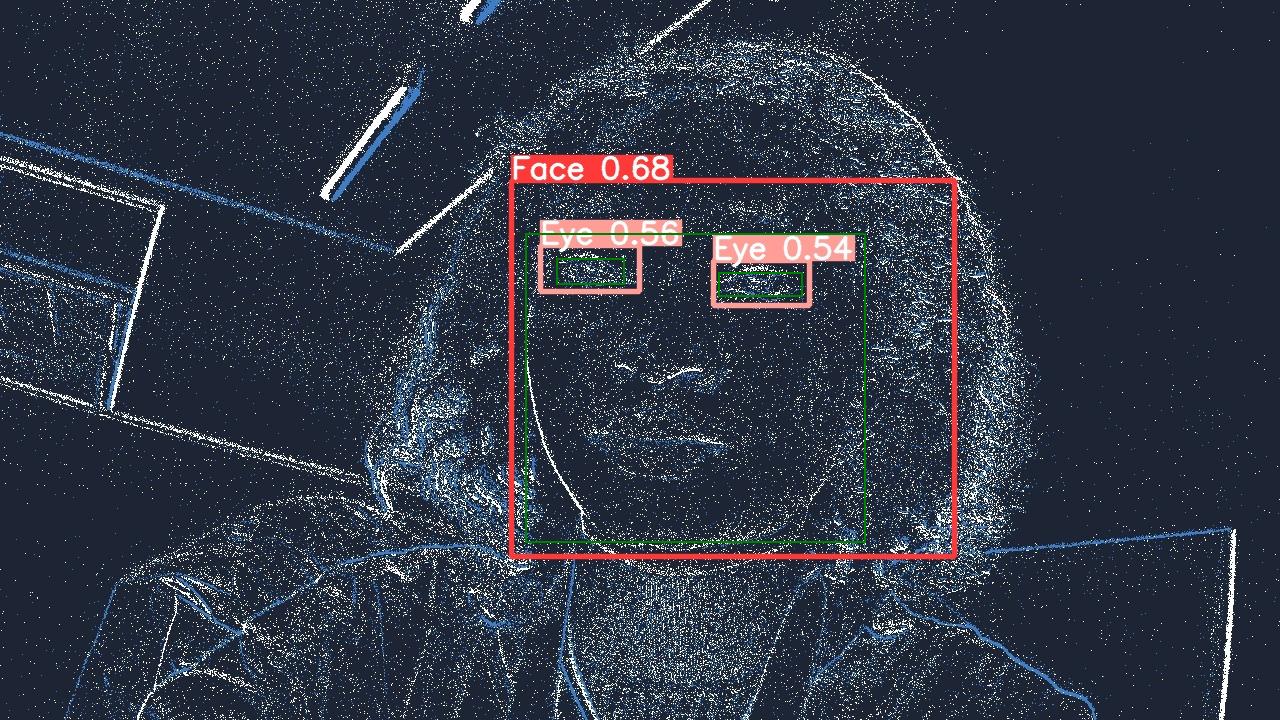}
		\caption{Local data}
		\label{fig5:a}
	\end{subfigure}
	\begin{subfigure}{0.3\linewidth}
		\includegraphics[width=3.5cm, height = 3cm]{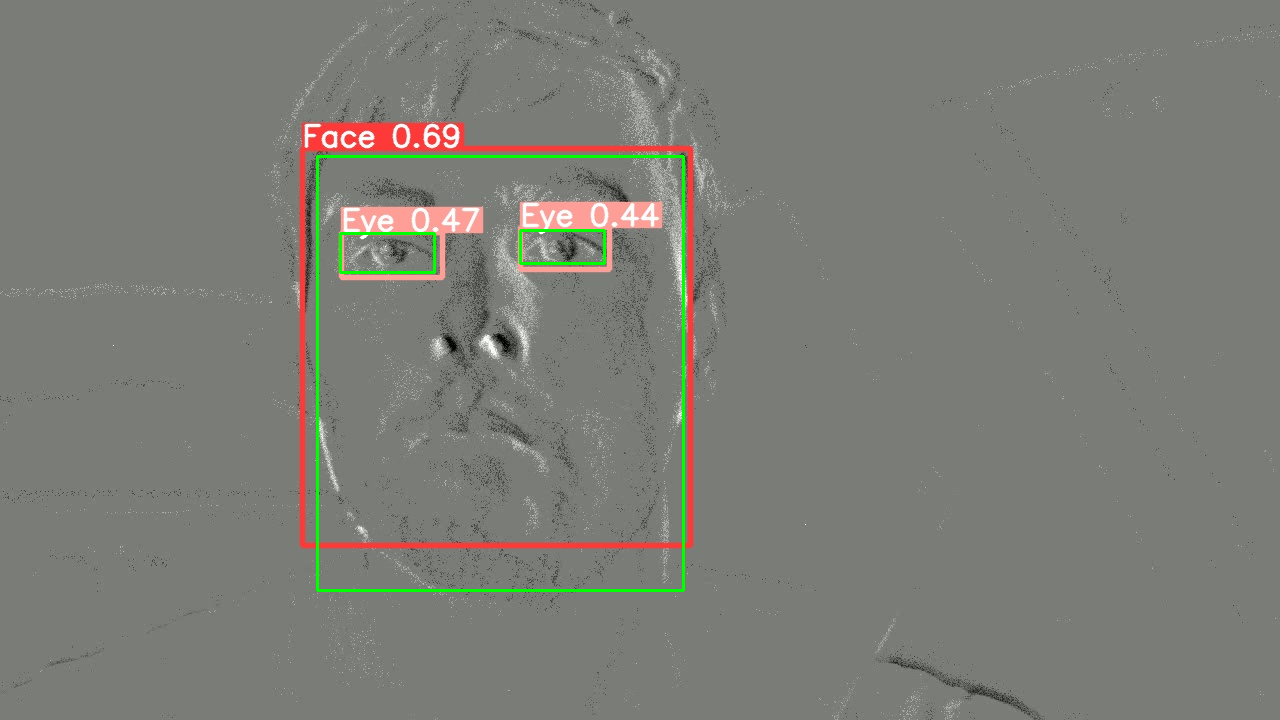}
		\caption{Data by Ryan et al ~\cite{Ryan2021}}
		\label{fig5:b}
	\end{subfigure}
	\begin{subfigure}{0.3\linewidth}
	        \includegraphics[width=3.5cm, height = 3cm]{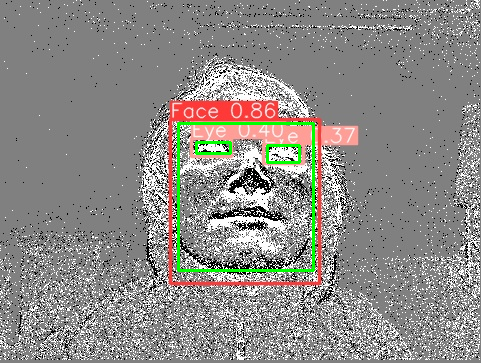}
	        \caption{FES data}
	        \label{fig5:c}
         \end{subfigure}
	\caption{Prediction Performance of YOLOv8.}
	\label{fig5:Test_V8}
\end{figure}

 The test videos used to record results qualitatively in this section include:  
 \begin{enumerate}
     \item Test video 1 recorded with Prophesee EVK4 in the lab with rapid head movement. Example frames shown in \cref{fig4:a} and \cref{fig5:a}
     \item Test video 2 from test subjects (Subject 0 test 1) used by~\cite{Ryan2021}, exhibiting slight head movements  with another fast-moving object within the field of view. Example frames shown in  \cref{fig4:b} and \cref{fig5:b}
     \item Test video 3 of subject 3 from FES dataset raw event streams with the subject wearing a nose mask. Example frames shown in \cref{fig4:c} and \cref{fig5:c}
 \end{enumerate}
 Test video 1 indicates performance in the presence of rapid motion, test video 2 indicates performance in different head positions while test video 3 indicates performance in the presence of occlusions. We use green boxes to indicate ground truth while purple boxes and red boxes in represent predictions of our models with confidence scores respectively.
 
The results presented show random frames from each video. Generally, this qualitative analysis highlights both models' capability of predicting faces and eyes from videos. We show results for 1 subject for each dataset and include full videos in supplementary materials at this url: \url{https://drive.google.com/drive/folders/1Wn1f1mpv5xqploAacKgsTnfqp5z2H6fY?usp=sharing}

%%%%%%%%%%%%%%%%%%%%%%%%%%%%%%%%%%%%%%%%%%%%%%%%%%%%%%%%%%%%%%%%%%%%%%%%%%%%%%%%%%%

\section{Conclusions}
This study uses a frame-based representation of events for face and eye tracking. The research addresses the challenge of under-sampling inherent in traditional RGB cameras and explores the advantages of using event cameras for face and eye tracking. By converting EC data into a format amenable to conventional deep learning algorithms, the study highlighted
the need for bridging the gap between the distinctive data format of ECS and the established paradigms of deep learning.  A testament to our methodological innovations is the successful generation of an event-based counterpart to the publicly available Helen Dataset, while optimizing the number of event frames produced during the simulation and thereby, enriching the resources available for face and eye tracking research. The efficacy of this dataset for face and eye tracking tasks is rigorously evaluated, and the findings are compared with those obtained using a voxel-based representation. The presented results affirm the dataset's reliability and its applicability to real-world event camera data. Our findings not only validate the dataset's utility but also highlight the distinct advantages of frame-based approaches, including its computational efficiency and accessibility. 

Looking forward, the application of event-based vision technology and event cameras holds potential for wide-ranging areas, including sports analytics, blink and saccade detection, emotion recognition, gaze tracking, \etc. The potential for applications in scenarios requiring adaptable lighting conditions, decrease of blur and adaptable resolution opens up new avenues for research and development.

% \section{Declaration of competing interest}
% We would like to  declare the following financial interests/personal relationships which may be considered as potential competing interests; This work is a result of a research collaboration between Xperi Galway and the Insight SFI centre for Data Analytics at the Dublin City University. 

\section{Acknowledgement}
This research was conducted with the financial support of Science Foundation Ireland (SFI) under grant no. [12/RC/2289\_P2] at the Insight SFI Research Centre for Data Analytics, Dublin City University in collaboration with FotoNation Ireland (Tobii).

% Now we have reached the maximum length of an ECCV \ECCVyear{} submission (excluding references).
% References should start immediately after the main text, but can continue past p.\ 14 if needed.
% \clearpage  % TODO REVIEW/FINAL: This \clearpage needs to be removed from both review and camera-ready versions.

% ---- Bibliography ----
%
% BibTeX users should specify bibliography style 'splncs04'.
% References will then be sorted and formatted in the correct style.
%
\bibliographystyle{splncs04}
\bibliography{egbib}
\end{document}